%% file: paper.tex
\documentclass{article}
\usepackage{kr}

\usepackage{times}
\usepackage{soul}
\usepackage{url}
\usepackage[hidelinks]{hyperref}
\usepackage[utf8]{inputenc}
\usepackage[small]{caption}
\usepackage{graphicx}
\usepackage{amsmath}
\usepackage{amsthm}
\usepackage{booktabs}
\usepackage{algorithm}
\usepackage{adjustbox}
\usepackage{algpseudocode}
\usepackage{amssymb}
\usepackage{enumerate}
\usepackage{pgfplots}
\usepackage{tablefootnote}
\usepackage{tikz}
\usepackage{wasysym}
\usepackage{xspace}
\usepackage{nicefrac}

\DeclareMathOperator*{\argmin}{argmin}

\newcommand{\mlp}{\mathbf{MLP}\xspace}
\newcommand{\logit}{\text{logit}\xspace}
\newcommand{\multiset}[1]{\{\!\!\{ #1 \}\!\!\}}
\newcommand{\agg}{\emph{agg}\xspace}
\newcommand{\reals}{\mathbb{R}}
\newcommand{\comb}{\emph{comb}\xspace}
\newcommand{\CC}{\ensuremath{\C_2}}
\newcommand{\CCC}{\ensuremath{\C_3}}
\newcommand{\citeay}[1]{\citeauthor{#1} (\citeyear{#1})}

\newcommand{\EXP}{\mathbb{E}}

\newcommand{\Omit}[1]{}
\newcommand{\tup}[1]{\langle #1 \rangle}

\usepackage[scr=boondox]{mathalfa}
\newcommand{\F}{\mathcal{F}}
\newcommand{\Q}{\mathcal{Q}}
\newcommand{\C}{\mathsf{C}}

\title{Learning General Policies with Policy Gradient Methods}

\author{%
Simon St\r{a}hlberg$^1$\and
Blai Bonet$^2$\and
Hector Geffner$^{3,1}$ \\
\affiliations
$^1$Link\"{o}ping University, Sweden\\
$^2$Universitat Pompeu Fabra, Spain\\
$^3$RWTH Aachen University, Germany\\
\emails
simon.stahlberg@liu.com,
bonetblai@gmail.com,
hector.geffner@ml.rwth-aachen.de
}

\begin{document}

\maketitle

\begin{abstract}
  \input{sections/abstract.tex}
\end{abstract}

\input{sections/introduction}
\input{sections/background}
\input{sections/rl}
\input{sections/general-rl}

\input{sections/neural_networks}
\input{sections/experiments}
\input{sections/related_work}

\input{sections/conclusions}
\input{sections/acknowledgements.tex}

\bibliographystyle{kr}
\bibliography{control}

\end{document}

%% file: sections/abstract.tex
While reinforcement learning methods have delivered remarkable results in a number of settings,
generalization,  i.e., the ability to produce policies that generalize in a reliable and systematic way,
has remained  a challenge. The problem of generalization has  been addressed formally  in
classical planning  where  provable correct  policies that generalize over all instances of a given domain
have been learned using combinatorial methods. The aim of this work is to bring these two research threads
together to illuminate the conditions under which (deep) reinforcement learning approaches,
and in particular, policy optimization methods, can be used to learn policies that generalize like combinatorial methods do.
We  draw on lessons learned from previous combinatorial and deep learning approaches, and extend them in a convenient way.
From the former, we model policies as  state transition classifiers,  as (ground)  actions are not general
and change from instance to instance. From the latter, we use graph neural networks (GNNs)  adapted to deal
with  relational structures for representing value functions over planning states, and in our case, policies.
With these ingredients in place, we find that  actor-critic methods can be used to  learn policies
that generalize almost as well as those obtained using combinatorial approaches while avoiding the
scalability bottleneck and  the use of feature pools. Moreover, the limitations of the DRL
methods on the benchmarks considered have little to do with deep learning or reinforcement learning algorithms,
and result from the well-understood expressive limitations of GNNs,
and the tradeoff between optimality and generalization  (general policies cannot be optimal in some domains).
Both of these limitations are addressed without changing the basic DRL methods by  adding
derived predicates and an alternative cost structure to optimize.


%% file: sections/introduction.tex
\section{Introduction}

Reinforcement learning (RL) has delivered remarkable results in a number
of settings like game playing and robotics \cite{dqn,drl:robotics,silver2}.
However, generalization, the ability to produce policies that generalize
in a reliable and systematic way, remains a challenge \cite{drl-generalization:survey}.
A basic question, for example, is whether RL methods can be used to learn policies that
generalize to any instance of a classical planning domain like Blockworld;
namely,  instances  that feature an arbitrary number of objects
that must be mapped into an arbitrary goal configuration. The power of deep learning for delivering such policies has been explored
in a number of works \cite{sylvie:asnet,mausam:dl,karpas:generalized};
yet these approaches do not result in nearly perfect general policies.

The problem of learning general policies has been addressed formally in the
KR and planning setting
\cite{srivastava08learning,hu:generalized,bonet:ijcai2015,BelleL16,bonet:ijcai2017,sheila:generalized2019}
and some  logical approaches appealing to combinatorial
optimization methods have been used to learn provable correct policies that generalize
over a number of classical planning domains \cite{frances:aaai2021,dominik:learning-sketches}.
Roughly, the (unsupervised) learning problem is cast as the joint problem of
selecting state features $f$ from a pool of features $\F$
and classify state  transitions $(s,s')$ into ``good'' and ``bad'', such that good
transitions lead to the goal with no cycles, and
good and bad transitions can be distinguished by their effects on the selected
features. The use of state transitions $(s,s')$ to select actions, rather than selecting
the actions directly, is because the set of (ground) actions  changes
with the set of objects, while the set of predicates, and hence, the features
derived from them, remain fixed \cite{martin:kr2000}.


The RL and planning approaches for learning general policies have different strengths
and weaknesses. The deep RL (DRL) approach does not offer a meaningful formal language to encode
general policies, nor a meaningful meta-language to study them, but its strengths are
that it does not require symbolic states (e.g., it can deal with states represented
by pixels), and it does not have a scalability bottleneck (stochastic gradient descent
is efficient and works surprisingly well).
On the other hand, the planning approach offers meaningful languages for representating instances
and domains, and  policies and states, while its shortcomings are its  reliance on
human-provided domain encodings,  and the computational bottleneck of the combinatorial
optimization methods: training instances must be large enough so that the resulting
policies generalize, and small enough so that the solvers can solve them optimally.

Works that combine  planning languages for representing states, and deep networks for
encoding either value and policy functions have brought the DRL and planning approaches closer.
\citeay{stahlberg:bonet:geffner:icaps:2022} use a graph neural network (GNN) for mapping planning states $s$ into a value function $V(s)$ that is trained,
in a supervised way, to approximate the optimal value function $V^*(s)$ that measures
the minimum number of steps  from $s$ to the goal. The learned value function $V$
encodes the greedy policy $\pi_V$ that selects successor states $s'$ with lowest $V(s')$
value. More recently, the same neural architecture has been  used to learn a value function $V$
without supervision by minimizing the Bellman errors $|V(s) - \min_{s'} [1+V(s')]|$, where $s'$ ranges
over the possible successors of state $s$ and the cost of actions is assumed to be $1$ \cite{stahlberg:bonet:geffner:kr:2022}.
The resulting method is a form of \emph{asynchronous value iteration} where the value
function $V$ is encoded by a deep net and the error is minimized at selected states $s$
by stochastic gradient descent.


These works, however, do not  answer the question of \emph{whether RL
methods can be used to learn policies that generalize in a systematic manner.}
We are particularly interested in \emph{policy gradient} approaches \cite{reinforce,sutton:book}
because of their broader scope. Unlike value-based methods,
they can be used 1)~when the state and action spaces are continuous, 2)~when the system
state is not fully observable, and also 3)~when the system state is not clearly defined,
as in ChatGPT \cite{chatgpt}. Additionally, expressing
a policy with features is often ``easier'' than expressing a value function with the same features
\cite{sutton:book,frances:ijcai2019,frances:aaai2021}.

Can policy gradient methods and in particular actor-critic algorithms be used to learn
nearly perfect policies for classical planning benchmark domains? And if so, how are
they to be used and what are their limitations?
The literature does not provide a crisp  answer to these questions, as the use of RL methods
has  focused on performance relative to baselines, and in classical planning, on learning
heuristics \cite{heuristics1,heuristics2,heuristics3}.
Approaches that have aimed at nearly perfect general policies have not relied on RL methods,
and much less on standard RL methods off the shelf. This is what we aim to do in this work.

The paper is organized as follows.
%
We first provide background on classical planning, dynamic programming (DP), and general policies.
Then, we review  RL algorithms as approximations of exact DP methods, and
adapt them to learn general policies. We  look then  at the representation of value and policy functions
in terms of deep networks, present and analyze the experimental results, discuss related work, and draw the conclusions.

%% file: sections/background.tex
\section{Background}

We review classical planning, the two basic dynamic programming methods, and general policies.

\subsection{Classical Planning}

A classical planning problem is a pair $P\,{=}\,\tup{D,I}$ where $D$ is a first-order
\emph{domain} and $I$ contains information about the instance \cite{geffner:book,ghallab:book,pddl:book}.
The domain $D$ has a set of predicate symbols $p$ and a set of action schemas with
preconditions and effects given by atoms $p(x_1, \ldots, x_k)$ where $p$ is a predicate
symbol of arity $k$, and each $x_i$ is an argument of the schema.
An instance is a tuple $I\,{=}\,\tup{O, \textit{Init},\textit{Goal}}$ where $O$ is a
set of object names $c_i$, and $\textit{Init}$ and $\textit{Goal}$ are sets of
\emph{ground atoms} $p(c_1, \ldots, c_k)$.

A classical problem $P\,{=}\,\tup{D,I}$ encodes a state model $S(P)\,{=}\,\tup{S,s_0,S_G,\textit{Act},A,f}$
in compact form where the states $s \in S$ are sets of ground atoms from $P$, $s_0$
is the initial state $I$, $S_G$ is the set of goal states $s$ such that $S_G \subseteq s$,
$\textit{Act}$ is the set of ground actions in $P$, $A(s)$ is the set of ground actions
whose preconditions are (true) in $s$, and $f$ is the induced transition function
where $f(a,s)$, for $a \in A(s)$, represents the state $s'$ that follows action $a$
in the state $s$.  An action sequence $a_0, \ldots, a_{n}$ is applicable in $P$ if
$a_i \in A(s_i)$ and $s_{i+1}\,{=}\,f(a_i,s_i)$, for $i\,{=}\,1,\ldots,n$, and it
is a plan if $s_{n+1} \in S_G$.
The \emph{cost} of a plan is  assumed to be given by its length and a plan is
\emph{optimal} if there is no shorter plan.

\subsection{Dynamic Programming}

In generalized planning one is interested in plans that reach the goal from any
state $s$ of a large collection of domain instances. Such plans can be represented indirectly
in compact form by means of value $V(s)$ or policy functions $\pi(s)$ that map
states into real numbers and actions respectively. Dynamic programming offers two
basic methods for computing such functions, value and policy iteration, in the more
general setting of Markov Decision Processes (MDPs) where the deterministic state
transition function $f(a,s)$ and uniform action costs are replaced by transition
probabilities $P_a(s'|s)$ and  costs $c(a,s)$, $a \in A(s)$. 


\medskip
\noindent\textbf{Value Iteration}
(VI) approximates the optimal value function $V^*(s)$, that provides the minimum
(discounted) cost to reach the goal from $s$, by computing an approximate solution
of the Bellman optimality equation
\begin{alignat}{1}
    \label{eq:bellman-opt}
    V(s)\ =\ \min_{a \in A(s)} \biggl[c(a,s) + \gamma \sum_{s'} P_a(s'|s) V(s') \biggr]
\end{alignat}
where $V(s)=0$ for goal states and $\gamma\in(0,1)$ is the discount factor.
This is done by initializing $V(s)$ to zero (although the initialization for non-goal
states can be arbitrary) and updating the value vector $V$ over non-goal
states $s$ as:
\begin{alignat}{1}
    \label{eq:full-update}
    V(s)\ :=\ \min_{a \in A(s)} \biggl[ c(a,s) + \gamma \sum_{s'} P_a(s'|s) V(s') \biggr] \,.
\end{alignat}

In VI, these updates are done in parallel by using two $V$-vectors, while in
\emph{asynchronous} VI, a single $V$-vector is used, and there is no requirement
that all states are updated at each iteration. In both cases, asymptotic
convergence to $V^*$ is guaranteed provided that all states are updated infinitely
often \cite{bertsekas:dp}. If $V$ is optimal, i.e., $V=V^*$, the policy $\pi_V$ that is greedy
in $V$, is also optimal, where:
\begin{alignat}{1}
    \label{eq:greedy}
    \pi_V(s)\ =\ \argmin_{a \in A(s)} \biggl[ c(a,s) + \gamma \sum_{s'} P_a(s'|s) V(s') \biggr] \,.
\end{alignat}

\Omit{
    A particular asynchronous VI algorithm, called RTDP, simulates executions of
    the greedy policy $\pi_V$ by iteratively sampling the successor states $s'$ of state $s$ with probability $P_a(s'|s)$ for $a=\pi(s)$.
    The states that are visited in these executions are then updated in sequence.
    Interestingly, if the initial value function $V=V_0$ is a lower bound, i.e., $V(s) \leq V^*(s)$ for all states,
    then $V(s)$ and $\pi_V(s)$ converge to the optimal values $V^*(s)$ and policy $\pi^*(s)$ over the states $s$
    that are reachable from the initial state $s_0$ with the optimal policy $\pi^*$. In other words, an ``optimistic''
    value function remains ``optimistic'' after full Bellman updates \eqref{eq:full-update}, and then no ``exploration''
    is needed and policies greedy in $V$ can be used. Otherwise, other states have to be updated too.
}

\medskip
\noindent\textbf{Policy Iteration}
(PI) iterates on policies $\pi$ rather than on value vectors $V$, and
consists of two steps: policy evaluation and policy improvement.
Starting with an arbitrary policy $\pi=\pi_0$, policy evaluation computes
the values $V^\pi(s)$ that encode the expected cost to the goal from $s$
when following policy $\pi$. These costs are computed by solving the
linear Bellman equation for policy $\pi$:
\begin{alignat}{1}
    \label{eq:bellman-pi-det}
    V(s)\ =\ \biggl[ c(a,s) + \gamma \sum_{s'} P_a(s'|s) V(s') \biggr]
\end{alignat}
for $a\,{=}\,\pi(s)$ and  $V(s)\,{=}\,0$ for goal states $s$.
If the policy $\pi$ is not  greedy  with respect to the
value function $V=V^\pi$, any policy $\pi'$ that is greedy with respect to
$V^\pi$ is strictly better than $\pi$; i.e., $V^{\pi'}(s) \leq V^\pi(s)$ for all
$s$ with the inequality being strict for some states.
Thus, in the policy improvement step, PI sets the policy $\pi$ to $\pi'$
and the process is repeated. This process finishes after a finite number
of iterations when a policy $\pi$ is obtained that is greedy relative to
the value function $V\,{=}\,V^\pi$, as such a a policy cannot be
improved and is optimal. The number of iterations is finite because
the number of deterministic policies $\pi$ that map states into actions
is finite too.

\subsection{General Policies}

Generalized planning studies the representation and computation of policies
that solve many classical planning instances from the same domain at once
\cite{srivastava08learning,hu:generalized,BelleL16,sheila:generalized2019}.
\Omit{
  For example, the class of problems $\Q$ may include all Blocksworld instances
where a given block $x$ must be cleared, or all instances of Blocksworld
for any goal. In the latter case, it is assumed then that atomic goals are
represented as part of the states so no explicit consideration of the goals
is needed (details below).
}
A critical issue is how to represent general policies.
Clearly, they cannot be represented by policies that map states into (ground) actions
because the number and name of the actions change with the set of objects.
One simple representation of general policies is in terms of general value functions
$V$ \cite{frances:ijcai2019,stahlberg:bonet:geffner:icaps:2022}.
These functions map states $s$ over the instances in a class $\Q$ into non-negative scalar values $V(s)$
that are zero only at goal states, and they can be used to define
greedy policies $\pi_V$ that select the actions that lead to successor
states $s'$ with minimum $V(s')$ value.
If the  value of the child $s'$  is always lower than the value of its parent
state $s$, the value function $V$ represents a general policy $\pi_V$ that
is guaranteed to solve any problem in the class  $\Q$.

Another  simple representation of general policies is as classifiers of
state transitions into two categories, good and  bad \cite{frances:aaai2021}.
If for each non-goal state $s$, there is a good state transition $(s,s')$,
and chains of good transitions do not lead to dead-end states or cycles,
the policy is guaranteed to solve any problem in the class.

A general policy is optimal if it results in shortest trajectories to the
goal, yet optimality for general policies is not necessary and it is often impossible.
For example, there are general policies for solving arbitraty instances
of Blocksworld, but there are no optimal general policies,\footnote{Policies
  that decide what action to apply in polynomial time.}
as optimal planning in Blocksworld, like in many other ``easy'' classical
planning domains, is intractable.

\Omit{ lack of space
The classifiers encoding general policies have been expressed and learned
in terms of pools of state features that are obtained from the primitive
domain predicates and a description logic grammars \cite{bonet:aaai2019}
while general value functions have been represented as functions of these
or other state features \cite{frances:ijcai2019},
or as GNNs that map  planning states into scalar
values \cite{stahlberg:bonet:geffner:kr:2022}. The two approaches are closely related as there is a known
correspondence between the state features that can be obtained from the
domain predicates using a description logic grammar and those that can be
learned by GNNs \cite{morris:gnn-wl,barcelo:gnn,grohe:gnn}.
}

In this work, we draw on these ideas to represent policies as state transition
classifiers using GNNs that accept state pairs $s$ and $s'$ and determine
whether state transitions $(s,s')$ are good or not. The policies are trained
without supervision using policy optimization algorithms, to be discussed next.

%% file: sections/rl.tex
\section{From Exact DP to Approximate RL}

While there are basically two exact model-based DP algorithms, there
are many model-free RL/DRL methods. Rather than jumping directly to the 
algorithms that we will use, it will be convenient to understand the latter
as suitable approximations of the former.

\subsection{Approximate Prediction: Learning $V$}

The main reason for doing approximation is that the number of states
$s$ may be too large or infinite, as in generalized planning.
We focus on approximating value functions $V^\pi$ since we aim
at policy optimization algorithms, yet similar ideas can be applied
for approximating the optimal value function  $V^*$.

For policy optimization algorithms, it is convenient to consider
\emph{stochastic policies} $\pi$ that assign probabilities $\pi(a|s)$
for selecting the action $a$ in state $s$, rather than deterministic
policies, as deep neural networks output real values.
The Bellman equation for evaluating a  stochastic policy $\pi$ is:
\begin{alignat}{1}
  \label{eq:bellman-pi-stoch}
  V(s)\ =\ \sum_{a} \pi(a|s) \biggl[ c(a,s) + \sum_{s'} P_a(s'|s) V(s') \biggr] \,.
\end{alignat}
The solution to this linear system of equations, provided that $V(s)\,{=}\,0$
for goal states, is $V\,{=}\,V^\pi$, and a common method to solve it is
by a form of value iteration that uses the Bellman equation for policy $\pi$ instead of the
optimality equation \eqref{eq:bellman-opt}, with updates of the form:
\begin{alignat}{1}
  \label{eq:bellman-pi-full}
  V(s) := \sum_{a} \pi(a|s) \biggl[ c(a,s) + \gamma \sum_{s'} P_a(s'|s) V(s') \biggr] \,.
\end{alignat}
\noindent Successive updates of this form ensure that $V$ eventually converges
to $V^\pi$.
Three common approximations of this policy evaluation and the resulting value function $V^\pi$ are:
\begin{enumerate}[$\bullet$]
  \item \textbf{Sampling of (seed) states.} As in asynchronous VI, states $s$ are selected
    for update at each iteration by some form of stochastic sampling which does not
    necessarily guarantee that all states are updated infinitely often.
  \item \textbf{Sampling of actions, costs, and successors.} Either because state transition
    probabilities and action costs are not known, or because there are too many actions
    or too many successor states, samples of the actions $a$, costs $c(a,s)$, and successor states $s'$
    are used instead of considering all actions and successors. The resulting
    \emph{sampled updated} are:
    \begin{alignat}{1}
      \label{eq:bellman-pi-sampled}
      V(S)\ :=\ V(S) + \alpha \bigl[ C + \gamma V(S') - V(S) \bigr] \,,
    \end{alignat}
    where $\alpha$ is the step size or \emph{learning rate}, is characteristic of
    stochastic approximation methods \cite{robbins-monro,stochastic-approx}.
    If the action $A$ is sampled with probability $\pi(A|S)$ and $S'$ with probability $P_A(S'|S)$,
    written $A\,{\sim}\,\pi(\cdot|S)$ and $S'\,{\sim}\,P_A(\cdot|S)$, the expression $C+\gamma V(S')$ 
    is an \emph{unbiased estimator} of the right-hand side of \eqref{eq:bellman-pi-full},
    and hence the sampled backups guarantee that $V$ converges to $V^\pi$ provided
    standard conditions on the step sizes and that all states $s$ are updated
    infinitely often.
  \item \textbf{Function approximation.} Finally, the value function $V(s)$ can be
    represented by a deep neural network with adjustable parameters $\omega$.
    The updates no longer change the value of entries in a table (tabular updates)
    but the value of the parameters $\omega$ for minimizing the \emph{squared loss}
    of the \emph{sampled Bellman residual}
    \begin{alignat*}{1}
      \frac{1}{2}\bigl[ C + \gamma V(S') - V(S) \bigr]^2
    \end{alignat*}
    through a step of gradient descent:
    \begin{alignat}{1}
      \label{V-update}
      \omega\ :=\ \omega + \alpha \bigl[ C + \gamma V(S') - V(S) \bigr] \nabla V(S) \,,
    \end{alignat}
    where $\alpha$ is the step size, $A\,{\sim}\,\pi(\cdot|S)$ and $S'\,{\sim}\,P_A(\cdot|S)$,
    and $\nabla V(S)$ is the gradient of the function $V$ relative to $\omega$ evaluated
    at the state $S$.
    This update expression follows from the standard formula for minimizing a function
    (locally) using its gradient except that it assumes that the ``target'' value $V(S')$
    does not depend on $\omega$, a so-called ``semi-gradient'' method \cite{sutton:book}.
\end{enumerate}

These three approximations are largely independent of each other and do not have to
be used together. For example, \citeay{rubik:drl}
learn an approximation of the optimal value function $V^*$ for
guiding the search in Rubik's Cube  by using value iteration
while representing the value function by a deep net.
A similar approach is used by \citeay{stahlberg:bonet:geffner:kr:2022}
for learning general value functions from small instances.
In the approach below, however, it is not the value functions that
``transfer'' (generalize) to new instances but the learned policy $\pi$ itself.

\subsection{Approximate Control: Learning $\pi$}

Policy gradient methods make use of approximations of the value functions $V^\pi$
for improving  a differentiable policy $\pi$ via gradient descent
\cite{reinforce,sutton:book}. If the expected cost $J(\pi)$ of a policy $\pi$ over an MDP with prior $h$ over the
initial states and parameters $\theta$ is
\begin{alignat}{1}
  J(\pi)\ =\ \textstyle \sum_{s} h(s) V^\pi(s) \,,
\end{alignat}
the gradient of $J(\pi)$ relative to the vector of parameters $\theta$
satisfies \cite[Policy Gradient Theorem]{sutton:book}:
\begin{alignat}{1}
  \label{policy-loss}
  \nabla J(\pi)\ &\propto\ \smash[b]{\sum_{s} \mu(s) \sum_a Q^\pi(s,a) \nabla \pi(a|s)} \\
  \intertext{where}
  Q^\pi(s,a)\ &=\ c(s,a) + \sum_{s'} P_a(s'|s) V^\pi(s')
\end{alignat}
and $\mu(s)$ stands for the fraction of times that executions of the policy $\pi$
visit the state $s$ when the prior is $h(s)$.
The sum in \eqref{policy-loss} corresponds to the expectation
\begin{alignat}{1}
  \label{eq:gradient}
  \nabla J(\pi)\ \propto\ \EXP_{S\sim\mu,A\sim\pi(\cdot|S)}\bigl[ Q^{\pi}(S,A) \nabla \ln\pi(A|S) \bigr]
\end{alignat}
where $S$ and $A$ are random variables that stand for a state and action sampled
according to $\mu$ and $\pi(\cdot|S)$ respectively.
Gradient descent adjusts the parameters $\theta$ of the policy as:
\begin{alignat}{1}
  \theta\ :=\ \theta - \alpha \nabla J(\pi) \,.
\end{alignat}
As before, common approximations of the gradient \eqref{eq:gradient} are:
\begin{enumerate}[$\bullet$]
  \item \textbf{Sampling states and actions.} Expectation in \eqref{eq:gradient}
    can be replaced by the unbiased estimator $Q^{\pi}(S,A) \nabla \ln\pi(A|S)$ where
    state $S$ and action $A$ are sampled by executing $\pi$.
  \item \textbf{Approximating $Q$- and $V$-values.} The $Q^{\pi}(S,A)$ value can be
    replaced by the unbiased estimator $C + V^\pi(S')$ with $S'\,{\sim}\,P_A(\cdot|S)$
    and $C\,{\sim}\,c(A,S)$,
    and $V^\pi(S')$ can be approximated as discussed above.
  \item \textbf{Use of baselines.} A term $b(S)$ is decremented from the $Q^{\pi}(S,A)$ value
    in \eqref{eq:gradient}, where $b(S)$ is a function that depends on the state $S$ but not
    on the action $A$ or the successor state $S'$. It can be shown that this offset does not
    affect the value of the expectation but reduces its variance \cite{actor-critic}.
\end{enumerate}

By combining these approximations,  the expression for the policy
parameter update becomes, for example,  one of the standard forms of the \emph{actor-critic (AC) RL algorithm}
\cite{actor-critic,sutton:book} where the policy parameters are updated as:
\begin{alignat}{1}
  \label{eq:gradient-update}
  \theta\ :=\ \theta - \alpha \bigl[ C + V(S') - V(S) \bigr] \nabla\ln\pi(A|S)
\end{alignat}
where $S$ and $A$ are sampled by executing the policy $\pi$, $C$ is the
sampled cost, $V$ is an approximation of $V^\pi(S)$, and the successor state
$S'$ is sampled with probability $S'\,{\sim}\,P_A(\cdot|S)$.
In this case, the baseline is $b(S)\,{=}\,V(S)$, and the same samples for
$S$, $A$ and $S'$ are used for updating $V$ using \eqref{V-update}.

Other AC variants can be obtained from other approximations of the
``actor's'' gradient $\nabla J(\pi)$ and the ``critic'' $V^\pi$.
Other RL algorithms are the so-called value-based methods, like SARSA and
Q-learning, that do not consider policies explicitly, and Monte-Carlo RL
methods like REINFORCE, that do not consider explicit value functions.

%% file: sections/general-rl.tex
\section{RL for Generalized Planning}

In our setting of generalized planning, states are planning states
represented as sets of ground atoms $p(t)$ where $p$ is a domain
predicate and $t$ is a tuple of objects of the same arity as $p$.
In addition, for generalizing to arbitrary conjunctive, ground goals,
the goals of a planning instance, which are given by ground atoms $p(t)$,
are represented as part of the state $s$ by using new predicate symbols $p_G$ \cite{martin:kr2000,bonet:aaai2019}.
Atoms like $clear(c)$ and $clear_G(c)$ in a state of Blocksworld, for example,
say that the block $c$ is clear in the state and that the block $c$ must be
clear in the goal, respectively.
A goal state in this representation is a state $s$ where for each goal atom
$p_G(t)$ in $s$, $p(t)$ is in $s$.

The two actor-critic algorithms for learning general policies over classical
planning domains using this representation of states are shown in
Figs.~\ref{alg:3:sampled:opt} and \ref{alg:3:full:opt}.

\begin{algorithm}[t]
  \begin{algorithmic}[1]
    \State \textbf{Input:} Training MDPs $\{M_i\}_i$, each with state priors $p_i$
    \State \textbf{Input:} Differentiable policy $\pi(s|s')$ with parameter $\theta$
    \State \textbf{Input:} Diff.\ value function $V(s)$ with parameter $\omega$
    \State \textbf{Parameters:} Step sizes $\alpha, \beta > 0$, discount factor $\gamma$
    \State Initialize parameters $\theta$ and $\omega$
    \State Loop forever:
    \State\quad Sample MDP index $i \in \{1, \dots, N\}$
    \State\quad\quad Sample non-goal state $S$ in $M_i$ with probability $p_i$
    \State\quad\quad Sample successor state $S'$ with probability $\pi(S'|S)$
    \State\quad\quad Let $\delta = 1 + \gamma V(S') - V(S)$
    \State\quad\quad  $\omega \gets \omega + \beta \delta \nabla V(S)$            \hfill Eq.~\eqref{eq:gradient-update}
    \State\quad\quad  $\theta \gets \theta - \alpha \delta \nabla \log \pi(S'|S)$ \hfill Eq.~\eqref{V-update}
    \State\quad\quad If $S'$ is a goal state,  $\omega \gets \omega - \beta V(S') \nabla V(S')$
  \end{algorithmic}
  \caption{Standard Actor-Critic for generalized planning: successor states $s'$ sampled with probability $\pi(s'|s)$.}
  \label{alg:3:sampled:opt}
\end{algorithm}

The first algorithm, AC-1, closely follows the gradient update rule in \eqref{eq:gradient-update} and the value update rule in \eqref{V-update}.
It is a standard AC algorithm where states $S$, actions $A$, and successor states $S'$ are sampled and adapted to the generalized setting.
\begin{enumerate}[$\bullet$]
  \item The stochastic policies $\pi$ are not assumed to encode a probability
    distribution over the actions in a given state because the set of actions
    changes from instance to instance.
    Policies are assumed instead to map states $s$ into a probability distribution
    $\pi(s'|s)$ over the set $N(s)$ of possible successor states $s'$ of $s$, and
    only indirectly, into a probability distribution over the actions applicable in $s$.
  \item The training set is given by a collection $\{M_i\}_i$ of
    (deterministic) MDPs. States $s$ are then sampled at training time from a sampled
    $M_i$ using a state prior $p_i$. The distributions  over
    states in $M_i$ and  over the training MDPs $M_i$ are assumed to
    be uniform.
  \item Action costs are all equal to 1 and thus there are no explicit sampled costs
    in the algorithm.
  \item When a sampled successor state $s'$ is a goal state, the value function parameters are updated to ensure that the value of the goal state is zero.
    This is achieved by minimizing the loss function $\nicefrac{1}{2}\,V(s')^2$, which drives the value at the goal state $s'$ towards zero.
  \item Finally, no trajectories are sampled; or equivalently, the only sampled
    trajectories have length $T\,{=}\,1$, corresponding to a single state transition.
    The reasons for this are discussed in the experimental section.

    \Omit{
    Namely, in each iteration, a single state $s$ is sampled using the state prior
    $p_i$, and from it, a successor $s'$ is sampled using the current policy, and the
    process repeats. Below we consider a more standard form of sampling where $T$
    consecutive transitions $(s_i,s_{i+1})$  from $s\,{=}\,s_0$ are sampled and used
    for updating the value and policy functions, before drawing another sample
    $s\,{=}\,s_0$.    This is indeed the form of sampling required by the expection in \eqref{eq:gradient},
    yet as we will see, it does not converge as well for learning a general policy, where
    all reachable states over all the instances can be initial states.
      }
\end{enumerate}

The second algorithm, AC-M in Fig.~\ref{alg:3:full:opt}, is a small variation of
the first where the seed states $S$ are sampled but not the successor states $S'$.
Instead, in the update expressions of the policy and value function parameters,
a sum over all possible successor states $s'$ is used, which in the case of
value updates, are weighted by the probabilities $\pi(s'|S)$ and result in full Bellman updates.
Algorithm AC-M is an actor-critic algorithm that makes use of the training models $M_i$
and which often converges faster than the model-free version AC-1.
Notice that for applying the learned policy $\pi$ from either AC-1 or AC-M, one must be
able to determine the possible successors $s'$ of a state $s$ in order to assess
the probabilities $\pi(s'|s)$ that define the stochastic policy.
A full model-free approach for generalized planning would need to learn this
structure as well.

\begin{algorithm}[t]
  \begin{algorithmic}[1]
    \State \textbf{Input:} Training MDPs $\{M_i\}_i$, each with state priors $p_i$
    \State \textbf{Input:} Differentiable policy $\pi(s|s')$ with parameter $\theta$
    \State \textbf{Input:} Diff.\ value function $V(s)$ with parameter $\omega$
    \State \textbf{Parameters:} Step sizes $\alpha, \beta > 0$, discount factor $\gamma$
    \State Initialize parameters $\theta$ and $\omega$
    \State Loop forever:
    \State\quad Sample MDP index $i \in \{1, \dots, N\}$
    \State\quad\quad Sample non-goal state $S$ in $M_i$ with probability $p_i$
    \State\quad\quad Let $V' = 1 + \gamma \Sigma_{s' \in N(S)} \left[ \pi(s'|S) V(s') \right]$
    \State\quad\quad Let $b(S) = V' - 1$ be the baseline
    \State\quad\quad $\omega \gets \omega + \beta (V' - V(S)) \nabla V(S)$
    \State\quad\quad $\theta \! \gets \!  \theta \!-\! \alpha \Sigma_{s' \in N(S)} \left[ (V(s') - b(S)) \nabla \pi(s'|S) \right]$
    \State\quad\quad If $s' \in N(S)$ is  goal state, $\omega \gets \omega - \beta V(s') \nabla V(s')$
  \end{algorithmic}
  \caption{All-Actions Actor-Critic: all successor states considered for updating policy and value functions.}
  \label{alg:3:full:opt}
\end{algorithm}

%% file: sections/neural_networks.tex
\section{Neural Network Architecture}

The value and policy functions, $V(s)$ and $\pi(s'|s)$, are represented using
graph neural networks \cite{gori:gnn,book:gnn} adapted for dealing with relational structures.
The GNNs produce object embeddings $\phi(o)$ that then feed suitable readout functions.
We adopt the GNN architecture of \citeauthor{stahlberg:bonet:geffner:icaps:2022} (\citeyear{stahlberg:bonet:geffner:icaps:2022,stahlberg:bonet:geffner:kr:2022})
for representing value functions over planning states. This architecture is a variation of a similar one used
for solving Max-CSP problems \cite{grohe:max-csp}. 

\subsection{GNNs on Graphs}

GNNs represent trainable, parametric, and generalizable functions over graphs
specified by means of aggregate and combination functions $\agg_i$
and $\comb_i$, and a readout function $F$.
The GNN maintains an embedding $f_i(v) \in \reals^k$ for each vertex $v$ of the input graph $G$.
Here, $i$ ranges from $0$ to $L$, which is the number of iterations or layers.
The vertex embeddings $f_0(v)$ are fixed and
the embeddings $f_{i+1}(v)$ for all $v$ are computed from the $f_i$ embeddings as:%
\begin{alignat}{1}
  \label{eq:msg-passing}
  f_{i+1}(v)\,{:=}\,\comb_i(f_i(v), \agg_i( \multiset{f_i(w) | w{\in}N_G(v) } ) )
\end{alignat}
where $N_G(v)$ is the set of neighbors for vertex $v$ in $G$, and
$\multiset{\ldots}$ denotes a multiset. This iteration is usually seen
as an exchange of messages among neighbor nodes in the graph.
Aggregation functions $\agg_i$ like max, sum, or smooth-max,
map arbitrary collections of real vectors of dimension $k$ into a single $\reals^k$ vector.
The combination functions $\comb_i$ map pairs of $\reals^k$ vectors
into a single $\reals^k$ vector. The embeddings $f_{L}(v)$
in the last layer are  aggregated and mapped into an
output by means of a readout function.  All the functions are
parametrized with weights that are learnable. By design,
the function computed by a GNN is \emph{invariant} with respect to graph isomorphisms,
and once a GNN is trained, its output is well defined for  graphs $G$ of any size.

\subsection{GNNs for Relational Structures}

\begin{algorithm}[t]
    \begin{algorithmic}[1]
      \State \textbf{Input:} State $s$ (set of atoms true in $s$), set of objects
      \State \textbf{Output:} Embeddings $f_L(o)$ for each object $o$
      \State $f_0(o) \sim \mathbf{0}^{k}$ for each object $o \in s$
      \State For $i \in \{0, \dots, L-1\}$
      \State\quad For each atom $q := p(o_1, \dots, o_m)$ true in state $s$:
      \State\quad\quad ${m}_{q,o}\ :=\ [\mlp_p(f_i(o_1), \ldots, f_i(o_m))]_j$
      \State\quad For each object $o$ in state $s$:
      \State\quad\quad $f_{i+1}(o)\ :=\ \mlp_U\bigl(f_i(o), \agg(\multiset{{m}_{q,o} | o\in q })\bigr)$
    \end{algorithmic}
  \caption{Graph Neural Network (GNN) architecture that maps state $s$ into object embedding $f(o)=f_L(o)$ from which value and policy functions defined.}
  \label{alg:architecture}
\end{algorithm}

States $s$ in planning do not represent graphs, but more general relational structures.
These structures are defined by a set of objects, a set of domain predicates, and the atoms $p(o_1,\ldots,o_m)$ that are true in the state.
The objects define the universe, the domain predicates define the relations, and the atoms represent their denotations.
The set of predicate symbols $p$ and their arities are fixed by the domain, but the sets of objects $o_i$ may change from instance to instance.
The GNN used for dealing with planning states $s$ computes object embeddings $f_i(o)$ for each of the objects $o$ in the input state $s$,
and rather than messages flowing from ``neighbor'' objects to objects as in \eqref{eq:msg-passing},
the messages flow from objects $o_i$ to the atoms $q$ in $s$ that include $o_i$, $q=p(o_1, \ldots, o_m)$, $1 \leq i \leq m$,
and from such atoms $q$ to all the objects $o_j$ involved in $q$ as (see Fig.~\ref{alg:architecture}):
\begin{alignat}{1}
  \label{eq:msg-passing:2}
  f_{i+1}(o)\,{:=}\,\comb_i (f_i(o), \agg_i( \multiset{m_{q,o} | o\in q, q \in s} ) )
\end{alignat}
where $m_{q,o}$ for $q=p(o_1,\ldots,o_m)$ and $o=o_j$ is:
\begin{alignat}{1}
  \label{eq:msg-atoms}
  {m}_{q,o}\ :=\ [\comb_p (f_i(o_1), \ldots, f_i(o_m))]_j \,.
\end{alignat}
In these updates, the combination function $\comb_i$ takes the concatenation of two real vectors of size $k$
and outputs a vector of size $k$, while the combination function $\comb_p$, that depends on the predicate symbol $p$,
takes the concatenation of $m$ vectors of size $k$, where $m$ is the arity of $p$,
and outputs $m$ vectors of size $k$ as well, one for each object involved in the $p$-atom.
The expression $[\ldots]_j$ in \eqref{eq:msg-atoms} selects the $j$-th such vector in the output.
In Figure~\ref{alg:architecture}, we use the same $\agg$ and $\comb$ in every layer, and their weights are shared.
Each of the multilayer perceptrons (MLPs) is composed of a residual block, followed by a linear layer that reshapes the output to the desired size.
A residual block consists of a linear layer, a non-linear activation function, and another linear layer.
The output of the residual block is the sum of the input and the result of the second linear layer.
We used the non-linear activation function \emph{Mish} in our implementation~\cite{misra:mish}.


\Omit{ Section is a bit long; I removed this .. but feel free ..
 The hyperparameter in the networks are the embedding dimension $k$
and the number of layers $L$. The initial embeddings $f_0(o)$
are obtained by concatenating a zero vector with a random
vector, each of dimension $k/2$, to break symmetries.
Key for the GNN to apply to any state over the domain is the use of
a single MLP$_p$ for each predicate symbol $p$ in the domain.
}

\subsection{From the Object Embeddings to $V$ and $\pi$}

The readout functions $V(s)$ and $\pi(s'|s)$ are computed by aggregating object embeddings.
If $f^s(o)\,{=}\,f_L(o)$ denotes the final embedding for object $o$ in state $s$, the value for $s$ is
\begin{alignat}{1}
  V(s)\ =\ \textstyle \mlp\bigl(\sum_{o \in O} f^s(o)\bigr) \,,
\end{alignat}
where the $\mlp$ outputs a single scalar.

The policy $\pi$, in turn, must yield the  probabilities $\pi(s'|s)$ for each
successor state $s'$ in $N(s)$. This is achieved by first computing \emph{logits} for
pairs $(s,s')$ and then passing the logits through a \emph{softmax}:
\begin{alignat}{1}
  \logit(s'|s) &=\ \textstyle\mlp\bigl(\sum_{o \in O} \mlp(f^s(o), f^{s'}(o))\bigr) \,, \\
  \pi(s'|s) &\propto\ \textstyle\exp\bigl(\logit(s'|s)\bigr)
\end{alignat}
where the inner $\mlp$ outputs a vector of size $2k$, and the outer $\mlp$ outputs
a single scalar. The purpose of the inner $\mlp$ is to derive new features that are
specific to the transition. For example, it can identify that the agent is no
longer holding an item. Note that we need to know all the successor states to
determine $\pi(s'|s)$ as the logits for all successors are needed to compute the softmax.

It is also important to note that the nets for $V(s)$ and $\pi(s'|s)$ share weights
since the object embeddings come from the same GNN. However, the weights used to
define the readout functions are not shared.

\Omit{
a probability distribution for the state $s$ and all its successor states in $N(s)$.
To begin, we compute the logit for a transition from $s$ to $s'$:
\begin{alignat}{1}
  \logit(s' \mid s) = \mlp\left(\sum_{o \in O} \mlp(f^s(o), f^{s'}(o))\right),
\end{alignat}
where the inner $\mlp$ outputs a vector of size $2k$, and the outer $\mlp$ outputs a single scalar.
The purpose of the inner $\mlp$ is to derive new features that are specific to the transition. For example, it can identify that the agent is no longer holding an item.
The logits are then used to create a probability distribution over $N(s)$ through the softmax function:
\begin{alignat}{1}
  \pi(s' \mid s) = \frac{e^{\logit(s'|s)}}{\sum_{s'' \in N(s)} e^{\logit(s'' \mid s)}}.
\end{alignat}
Note that we need to know all the successor states to determine the transition probability from one state to another.
It is also important to note that the $V(s)$ and $\pi(s \mid s)$ share weights since the object embeddings come from the same GNN. However, the weights used to define the readout functions are not shared.

Finally, to define a value or policy function in relation to a goal, we add \emph{goal atoms} to the state $s$, which are defined using \emph{goal predicates}.
Specifically, we define a new predicate $P_G$ for each predicate $P$, and we add the atom $p_G(o_1, \dots, o_n)$ to the input state of the GNN for each atom $p(o_1, \dots, o_n)$ in the goal.
}

%% file: sections/experiments.tex
\section{Experimental Results}

The experiments test the generalization, coverage, and quality of the plans obtained by the learned policies.
We describe the data for training and testing, and the results obtained.
We aim at crisp results that mean close to $100\%$ generalization, and when this is not possible,
to provide clear explanations and in some cases
logical fixes that restore generalization.

\subsubsection{Data}
To create the training and validation sets, we generate the reachable state space from the initial state, along with shortest paths to a goal state.
We limit the training set to small instances with a maximum of 200,000 transitions.
If this condition is not met by the IPC instances or if better diversity is needed, we generate our own instances, aiming for approximately 24 per domain in the test set.

We used the same domains as \citeay{stahlberg:bonet:geffner:kr:2022} in our experiments, with the addition of Grid.
The domains and the data used for each are:
\begin{enumerate}[$\bullet$]\small
  \item \textbf{Blocks.}
    The goal is to build a single tower but in Blocks-multiple the goal consists of multiple towers.
    The training, validation and test sets have instances from the IPC with 4-7, 7, and 8-17 blocks, resp.
    For Blocks-multiple, the test set includes instances with up to 20 blocks.
  \item \textbf{Delivery.}
    The problem involves picking up objects in a grid with no obstacles and delivering
    them one by one to a target cell.
    The instances consist of up to $9\,{\times}\,9$ grids with up to $4$ packages.
    The training, validation, and test sets are partitioned based on the size of the
    reachable state space. The training instances are smaller than the validation instances,
    and the largest are in the test set.
  \item \textbf{Grid.}
    The goal is to find keys, open locked doors, and place specific keys at certain locations.
    The instances consist of up to $9\,{\times}\,7$ grids with up to $3$ locks and $5$ keys.
    The instances are partitioned as in Delivery into training, validation and test set.
  \item \textbf{Gripper.}
    The task is to move balls from one room to another using a robot with two grippers.
    The training, validation, and test instances contain of 1-9, 10, and 12-50 balls,
    respectively.
    The IPC instances only go up to 42 balls.
  \item \textbf{Logistics.}
    Domain that involves packages, cities, trucks and airplanes.
    The instances vary between 1-2 airplanes, 1-4 cities with 2 locations each, and 1-6
    packages, with exactly one truck in each city.
    Instances are partitioned into training, validation and test set on the size of rechable state space.
  \item \textbf{Miconic.}
    Planning for lift that picks and delivers passengers at different floors.
    The number of floors and passengers for training and validation vary between 2-8 and 1-5, resp.
    The instances with the largest state space are for validation.
    For the test set, we use IPC instances that feature up to $59$ floors and $29$ people
  \item \textbf{Reward.}
    Move in a grid with obstacles to pick up all rewards.
    The instances are from \citeay{frances:aaai2021}.
    The training, validation, and test sets consist of square grids with widths in 3-10, 10, and 15-25, resp.
    The maximum number of rewards for training and validation is $8$, and $23$ for testing.
  \item \textbf{Spanner.}
    The task is to tighten nuts at one end of a corridor with spanners that need to be collected on the way.
    The number of locations varies between 1-20 for training, validation, and testing, with at most $4$
    nuts for training and validation, and $12$ nuts for testing.
    The training and validation sets have no more than $6$ spanners, while the test set have up to $24$.
  \item \textbf{Visitall.}
    The task is to visit all or some cells in a grid with no obstacles.
    The grid sizes for training are limited to $3\,{\times}\,3$, $4\,{\times}\,2$, and smaller.
    The validation set uses $5\,{\times}\,2$ grids, while the test set includes
    grids of sizes ranging from $4\,{\times}\,4$ to $10\,{\times}\,10$.
\end{enumerate}

\subsubsection{Setup}
The GNN architecture is instantiated with hyperparameters $k\,{=}\,64$ and $L\,{=}\,30$,
and the discount factor $\gamma\,{=}\,0.999$ is used in the AC algorithms.
The hyperparameters $k$ and $L$ affect training speed, memory usage, and generalization (e.g., the GNN cannot compute shortest paths of length longer than $L$).
The architecture is implemented in PyTorch~\cite{pytorch} using Adam with learning rate of $0.0002$ \cite{kingma15}.\footnote{Code and data: \url{https://zenodo.org/record/7993858}}
The networks are trained using NVIDIA A100 GPUs for up to 6 hours.
Two models for each domain are trained, and the final model is the one with the best policy evaluation average on the validation set.
The quality of the plans is determined by comparing their length to the length of optimal plans computed with
Fast Downward (FD) \cite{helmert:fd} using the \emph{seq-opt-merge-and-shrink} portfolio with a time limit of
$10$ minutes and $64$ GB of memory on a Ryzen 9 5900X CPU.

\subsubsection{Results}

\begin{table*}[t]
  \centering
  \footnotesize
  \setlength{\tabcolsep}{2.25pt}
  \input{sections/tables/coverage.tex}
  \caption{%
    Performance of learned policies.
    The top subtable displays the results obtained from using standard update rules for Actor-Critic, while the middle and bottom subtables display the results obtained from using all-actions update rules for Actor-Critic, where the critic is either a neural network or a table, respectively.
    The learned policies were tested as both stochastic policies and deterministic policies.
    In the former, a transition is selected with the probability given by the policy, while in the latter, the most likely transition is always selected and
    the set of visited states is tracked to prevent cycles.
    The domains are shown on the left along with the number of instances in the test set.
    The coverage refers to the number of solved problems, and each policy was given a maximum of $10.000$ steps to reach a goal state.
    L represents the sum of the solution lengths over the test instances solved by the learned policy.
    PQ is a measure of overall plan quality given by the ratio of the sum of the plan lengths found by the learned policy and the optimal policy, which was determined with the help of Fast-Downward (FD).
    The number within parenthesis represents the number of instances FD was able to solve within our time and memory constraints.
    The columns $V_{\pi^*}$ and $V_{\pi}$ display the average value over the states in the validation set for the optimal policy and the learned policy, respectively,
    determined by policy evaluation using the stochastic policies.
  }
  \label{tbl:experiments:coverage}
\end{table*}

We tested the learned policies in two modes.
The first as a \emph{stochastic policy} that selects actions randomly following the distribution provided by the policy.
The second as a \emph{deterministic policy} that selects the most probable successor state.
Stochastic policies have the advantage of being able to escape cycles eventually. 
However, this is not possible with deterministic policies, so we used a closed set to avoid sampling an already visited successor.
Executions are terminated when the goal is not reached within 10,000 steps.

Table~\ref{tbl:experiments:coverage} presents the experimental results, divided in two subtables:
learning using the standard, sampled Actor-Critic algorithm (top), and learning using the all-actions Actor-Critic (bottom).
The left side of each subtable shows the results obtained with the stochastic policy, while the right hand side shows
the results obtained with the deterministic policy. 
We discuss test coverage and analyze the limitations encountered, which as we will see, \emph{have more to do with representation and complexity
  issues than with reinforcement or deep learning}. Indeed, we will use the analysis to address
the limitations in an informed manner. Later, we discuss plan quality and the impact of sampled trajectory
length on learning, where the result in the table are all for length 1. 

\subsubsection{Coverage}
We got a very high coverage out of the box in 6 out of the 10 domains considered, where the learned policy either solves all instances or almost all instances.
It is surprising, however, that one of the domains where we got perfect coverage for is Blocks.
This is unexpected because finding an optimal solution for Blocks instances is an NP-hard problem, thus there does not exists a compact policy.
In the IPC problems, the goal is to build a single tower.
We tested whether it is possible to learn a policy that does not rely on this goal structure and can construct any number of towers instead.
We achieved perfect coverage for this version, and the results are not due to "luck" because each successor state was selected with $100 \%$ probability (rounded to the nearest integer).
As these policies were learned using actor-critic, we can also define a greedy policy based on the learned value function (critic), which selects the successor state with the lowest value.
These policies solved all but three Blocks problems with eight blocks, for the model learned using all-actions Actor-Critic, and in these cases, the critic and the actor disagreed on the best successor state.

\subsubsection{Quality}

The last two columns in Table~\ref{tbl:experiments:coverage} show the average (policy evaluation) value over the entire validation set for the optimal policy and the learned policy.
This value represents the expected number of steps to a goal state from a uniformly sampled initial state.
When the difference between $V^{\pi}$ and $V^*$ is small, we can expect the policy to generalize well and produce plans that are close to optimal.


\subsubsection{Sampling and Length of Sampled Trajectories}

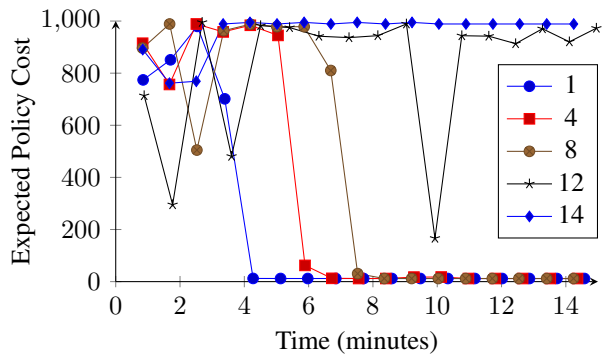
\begin{figure}[t]
  \centering
  \input{sections/curves/horizons.tex}
  \vskip -.5em
  \caption{Expected policy cost over the validation set during training sessions for Gripper with standard Actor-Critic for different
    trajectory lengths $T=1, 4, 8, 12, 14$. Table of results used $T=1$ which as seen here, does best.
    The training set included instances with up to 7 balls, while the validation instance had 8 balls.}
  \label{fig:horizons}
\end{figure}

Algorithms AC-1 and AC-M implicitly assume sampled trajectories of length 1.
After sampling a state $s$ from a training MDP with a uniform prior, a successor state $s'$ is sampled using probability $\pi(s'|s)$, and the process continues with another state $s$ sampled uniformly.
In RL practice, longer trajectories are commonly sampled, involving updating value and policy functions at each state before sampling another initial state.
In our setting, where we seek general policies and there are no privileged initial states, this alternative strategy was found to be detrimental to performance, as shown in Figure~\ref{fig:horizons}, and also in vanilla and tuned implementations of PPO that we tried \cite{schulman:ppo,stable-baselines3}.
Sampling state transitions from a buffer of stored experiences is common in the so-called experience replay and off-policy methods \cite{dqn,fujimoto:td3,haarnoja:sac}.

\subsubsection{Identifying and Overcoming Limitations}

Four domains, namely Grid, Logistics, Reward, and Spanner, were not solved nearly perfectly in any mode.
Interestingly, failures can be more instructive than successes, and this is no exception.
Below, we analyze the causes of these failures and the ways to address them.
Methodologically, this is important, as the first instinct when an RL algorithm fails is to try a different one.
This may improve the numbers, but in this case, as we will see, these changes will not be sufficient or necessary.
We will demonstrate how to achieve perfect coverage in these domains by sticking to our basic RL algorithms.
The limitations do not stem from RL algorithms, but from logical and complexity considerations.

The performance in these four domains results is affected by the  \emph{limited expressivity of GNNs} and
the \emph{optimality/generality tradeoff} \cite{stahlberg:bonet:geffner:kr:2022}.
GNNs can capture the  features that can be expressed in the two-variable fragment of first-order logic
with counting, $\CC$, but not those in larger fragments like $\CCC$ \cite{barcelo:gnn,grohe:gnn},
that are needed in Grid and Logistics.
In addition, the number of layers $L$ in the GNN
puts a limit on the lengths of the distances that can be computed; a problem
that surfaces in the Reward domain, where  the learned ``agent''cannot
move to the closest reward because it is just too far away.
The optimality/generalization tradeoff arises in domains like Logistics
and Grid that admit compact general policies but no compact policies
that are optimal, because optimal planning in both is NP-hard \cite{helmert:aij:2003}.
Yet RL algorithms aim to learn optimal policies which thus cannot generalize properly.

In Logistics, the GNN cannot determine whether a vehicle carrying a package is located in the city where the package needs to be delivered
as this requires $\CCC$ expressiveness. For learning a policy with the same GNN architecture, the states are extended to include (derived) atoms
indicating whether the package is in the correct city, regardless of whether it is on a plane, a truck, or at an incorrect location within the same city.
This version of Logistics achieved $91 \%$ coverage (with a plan quality of $4.53 = 1612 / 356$ $(18)$), which we trained using an all-action Actor-Critic
and evaluated as a deterministic policy.
%
%
For addressing, the optimality/generalization tradeoff, we modified the cost structure of the training MDPs
so that the learned policies optimize the probability of reaching the goal without entering a cycle
instead of the expected cost.\footnote{
  We omit the details of the resulting algorithm for lack of space,
  but it involves a third parametric function $D$ that  approximates this  probability.
  The policy $\pi$ optimizes $D^\pi$ and uses $V$  to avoid cycles (the only possible
   successors of $s$ for evaluating $D^\pi$ are those that decrease $V$ by more than
   $\epsilon$. In the experiments, $\epsilon = 0.75$).}
\Omit{
Perfect coverage cannot be expected in either Logistics or Grid due to the optimality/generalization
tradeoff mentioned above (this tradeoff affects Blocks too but suprisingly it did not affect coverage).
For this, the cost structure of the All-Actions Actor-Critic algorithm is changed so that the algorithm
(Algorithm~\ref{alg:3:full:subopt}) does not look for a policy $\pi$ that minimizes the expected cost $V^\pi$ but one that maximizes
the probability $D^\pi$ of reaching the goal without cycling (actions not chosen greedily in $V$ but must decrease $V$ by more
than $\epsilon > 0$, where $\epsilon = 0.75$ in the experiments).
}
Provided with the extra predicate and the new cost structure, a coverage of $100 \%$
was achieved in Logistics with a plan quality of $1.11 = 410/368$ $(19)$ using a deterministic policy
(we expect to obtain similar results in Grid with these two  extensions, but we have not achieved them yet
because the instance generator produces  too many unsolvable or trivial instances).


The final domain without full coverage is Spanner (Table~\ref{tbl:experiments:coverage}), where we observed
that the learned policies  fail when given instances with more spanners or nuts than those in the training instances.
There is no logical reason for this failure since dead-end state detection is possible with $\CC$ features~\cite{stahlberg-et-al-ijcai2021}.
To tackle this issue, we tested a \emph{tabular version} of the all-actions Actor-Critic algorithm
where the learned transition probabilities $\pi(s' \mid s)$ are stored and then used to solve the linear Bellman equation for $V^\pi$.
This approach gives accurate values, which for dead-end states are \emph{always} $\frac{1}{1 - \gamma}$
where $\gamma$ is the discount factor. 
The learned  policy  achieved then  $100 \%$ coverage in both modes (stochastic and deterministic) with a plan quality of $1.09 = 315 / 290$ $(10)$,
suggesting that the problem in the two Actor-Critic algorithms is that they do  not  sample dead-end states enough.
The tabular  evaluation of  policies, however, can only deal with small instances, and for this reason it only worked
in one other domain, Visitall.

%% file: sections/tables/coverage.tex
\begin{tabular}{@{\extracolsep{3.1pt}}llrllrlrr@{}}
    \toprule
                         & \multicolumn{3}{c}{Stochastic Policy}                 & \multicolumn{3}{c}{Deterministic Policy} & \multicolumn{2}{c}{Validation}                                                                                 \\ \cmidrule{2-4} \cmidrule{5-7} \cmidrule{8-9}
    Domain (\#)          & Coverage (\%)                                         & L                                        & PQ = PL / OL (\#)              & Coverage (\%) & L      & PQ = PL / OL (\#)        & $V_{\pi^{*}}$ & $V_{\pi}$ \\
    \midrule
                         &                                                       &                                          &                                &               &        &                          &               &           \\
    [-.5em]
                         & \multicolumn{6}{c}{\textbf{Standard Actor-Critic}}    &                                          &                                                                                                                \\
    \midrule
    \midrule
    Blocks (23)          & 23 (100 \%)                                           & 810                                      & 1.00 = 478 / 476 (16)          & 23 (100 \%)   & 810    & 1.00 = 478 / 476 (16)    & 18.60         & 18.60     \\
    Blocks-multiple (26) & 26 (100 \%)                                           & 898                                      & 1.00 = 450 / 448 (16)          & 26 (100 \%)   & 898    & 1.00 = 450 / 448 (16)    & 17.57         & 17.57     \\
    Delivery (24)        & 21 (88 \%)                                            & 591                                      & 1.01 = 544 / 540 (20)          & 23 (96 \%)    & 701    & 1.02 = 596 / 586 (21)    & 16.23         & 34.46     \\
    Grid (14)            & 5 (36 \%)                                             & 43                                       & 1.00 = 43 / 43 (5)             & 10 (71 \%)    & 516    & 2.68 = 356 / 133 (9)     & 18.13         & 668.28    \\
    Gripper (20)         & 17 (85 \%)                                            & 5~031                                    & 1.00 = 176 / 176 (4)           & 16 (80 \%)    & 1~312  & 1.00 = 176 / 176 (4)     & 14.89         & 14.89     \\
    Logistics (22)       & 3 (14 \%)                                             & 33                                       & 1.10 = 33 / 30 (3)             & 17 (77 \%)    & 37~667 & 77.4 = 23~839 / 308 (15) & 9.04          & 636.43    \\
    Miconic (30)         & 29 (97 \%)                                            & 1~438                                    & 1.00 = 185 / 185 (10)          & 29 (97 \%)    & 1~438  & 1.00 = 185 / 185 (10)    & 6.42          & 6.42      \\
    Reward (15)          & 13 (87 \%)                                            & 2~953                                    & 2.38 = 1~328 / 558 (7)         & 6 (40 \%)     & 655    & 1.24 = 362 / 292 (4)     & 22.15         & 120.34    \\
    Spanner (22)         & 18 (82 \%)                                            & 682                                      & 1.11 = 218 / 197 (7)           & 18 (82 \%)    & 678    & 1.11 = 218 / 197 (7)     & 130.74        & 130.82    \\
    Visitall (24)        & 24 (100 \%)                                           & 1~083                                    & 1.23 = 762 / 621 (20)          & 24 (100 \%)   & 1~032  & 1.12 = 696 / 621 (20)    & 4.36          & 4.39      \\
    \midrule
    Total (220)          & 179 (81 \%)                                           & 13~562                                   & -                              & 192 (87 \%)   & 45~707 & -                        & -             & -         \\
                         &                                                       &                                          &                                &               &        &                          &               &           \\
    [-.5em]
                         & \multicolumn{6}{c}{\textbf{All-Actions Actor-Critic}} &                                          &                                                                                                                \\
    \midrule
    \midrule
    Blocks (23)          & 23 (100 \%)                                           & 806                                      & 1.00 = 476 / 476 (16)          & 23 (100 \%)   & 806    & 1.00 = 476 / 476 (16)    & 18.60         & 18.60     \\
    Blocks-multiple (26) & 26 (100 \%)                                           & 902                                      & 1.00 = 450 / 448 (16)          & 26 (100 \%)   & 902    & 1.00 = 450 / 448 (16)    & 17.57         & 17.57     \\
    Delivery (24)        & 23 (96 \%)                                            & 691                                      & 1.00 = 586 / 586 (21)          & 24 (100 \%)   & 757    & 1.03 = 652 / 632 (22)    & 16.23         & 16.23     \\
    Grid (14)            & 6 (43 \%)                                             & 71                                       & 1.00 = 71 / 71 (6)             & 11 (79 \%)    & 292    & 1.50 = 248 / 165 (10)    & 18.13         & 84.21     \\
    Gripper (20)         & 20 (100 \%)                                           & 1~840                                    & 1.00 = 176 / 176 (4)           & 20 (100 \%)   & 1~840  & 1.00 = 176 / 176 (4)     & 14.89         & 14.89     \\
    Logistics (22)       & 0 (0 \%)                                              & -                                        & -                              & 8 (36 \%)     & 7~981  & 63.8 = 7~981 / 125 (8)   & 9.04          & 511.76    \\
    Miconic (30)         & 30 (100 \%)                                           & 1~527                                    & 1.00 = 185 / 185 (10)          & 30 (100 \%)   & 1~527  & 1.00 = 185 / 185 (10)    & 6.42          & 6.42      \\
    Reward (15)          & 7 (47 \%)                                             & 2~493                                    & 1.29 = 442 / 342 (4)           & 12 (80 \%)    & 1~464  & 1.21 = 582 / 481 (6)     & 22.15         & 167.25    \\
    Spanner (22)         & 15 (68 \%)                                            & 552                                      & 1.10 = 149 / 135 (5)           & 15 (68 \%)    & 552    & 1.10 = 149 / 135 (5)     & 130.74        & 130.81    \\
    Visitall (24)        & 23 (96 \%)                                            & 5~670                                    & 7.05 = 3~934 / 558 (19)        & 24 (100 \%)   & 971    & 1.08 = 671 / 621 (20)    & 4.36          & 4.40      \\
    \midrule
    Total (220)          & 173 (79 \%)                                           & 14~552                                   & -                              & 193 (88 \%)   & 17~092 & -                        & -             & -         \\
    \bottomrule
\end{tabular}

%% file: sections/curves/horizons.tex
\begin{tikzpicture}
    \begin{axis}[
        width=0.95\columnwidth,
        height=0.60\columnwidth,
        axis x line=bottom,
        axis y line=left,
        xmin=0,
        xmax=15,
        ymin=0,
        ymax=1000,
        legend style={at={(1.0, 0.5)}, anchor=east},
        xlabel=Time (minutes),
        ylabel=Expected Policy Cost
    ]
      \addlegendentry{1}
      \addplot coordinates {
        (0.85200, 774.17817)
        (1.70350, 851.24689)
        (2.53750, 979.54230)
        (3.39167, 701.49100)
        (4.27317, 12.23141)
        (5.12733, 11.98703)
        (5.97283, 11.96825)
        (6.84517, 11.96944)
        (7.71067, 11.99436)
        (8.59017, 11.93853)
        (9.47500, 11.93784)
        (10.32950, 11.93740)
        (11.17117, 11.93671)
        (12.02050, 11.93633)
        (12.86683, 11.93579)
        (13.72300, 11.93880)
        (14.57917, 11.94175)
      };
      \addlegendentry{4}
      \addplot coordinates {
        (0.83733, 914.92562)
        (1.67133, 756.62961)
        (2.50750, 988.82276)
        (3.34233, 958.36011)
        (4.18633, 983.83989)
        (5.04683, 944.96033)
        (5.88767, 62.29790)
        (6.72167, 12.64728)
        (7.55617, 11.92702)
        (8.40117, 11.94331)
        (9.27533, 17.34748)
        (10.12350, 17.65943)
        (10.97083, 11.92700)
        (11.80717, 11.92696)
        (12.65533, 11.92695)
        (13.51683, 11.92695)
        (14.36267, 11.92695)
      };
      \addlegendentry{8}
      \addplot coordinates {
        (0.83517, 896.11605)
        (1.67600, 988.76948)
        (2.52383, 504.86182)
        (3.36033, 963.07751)
        (4.18867, 988.04127)
        (5.01100, 978.44795)
        (5.86167, 978.29097)
        (6.68917, 810.27590)
        (7.52483, 30.69827)
        (8.35300, 11.93478)
        (9.20333, 11.92714)
        (10.03533, 11.92696)
        (10.88233, 11.92698)
        (11.71233, 11.92697)
        (12.53933, 11.92696)
        (13.40383, 11.92697)
        (14.23817, 11.92695)
      };
      \addlegendentry{12}
      \addplot coordinates {
        (0.88467, 713.62549)
        (1.76933, 295.23358)
        (2.68817, 994.86030)
        (3.60317, 481.00195)
        (4.50600, 982.35381)
        (5.40767, 975.00687)
        (6.31950, 941.83161)
        (7.25617, 936.40342)
        (8.15250, 944.02587)
        (9.04150, 988.78725)
        (9.92917, 166.42253)
        (10.76433, 943.77037)
        (11.60367, 941.98730)
        (12.43800, 913.24004)
        (13.27317, 970.54365)
        (14.10983, 920.33282)
        (14.95000, 972.66141)
      };
      \addlegendentry{14}
      \addplot coordinates {
        (0.84083, 890.63934)
        (1.67117, 761.56197)
        (2.50567, 768.96506)
        (3.34100, 988.78753)
        (4.17500, 994.86012)
        (5.01667, 988.78725)
        (5.85933, 994.86304)
        (6.69183, 988.78725)
        (7.50967, 994.84621)
        (8.39067, 988.78726)
        (9.21600, 994.87631)
        (10.05083, 988.78725)
        (10.89317, 988.80285)
        (11.73333, 988.78725)
        (12.56500, 988.78725)
        (13.40550, 988.78725)
        (14.24633, 988.75155)
      };
    \end{axis}
  \end{tikzpicture}

%% file: sections/related_work.tex
\section{Related Work}

\subsubsection*{General  Policies Using Deep (Reinforcement) Learning} DL and DRL methods \cite{sutton:book,bertsekas:dp,drl-book}
have been used to learn general policies \cite{drl-generalization:survey}.  In some cases, the  planning representation of the domains
is used  \cite{sylvie:asnet,mausam:dl,karpas:generalized};
in most  cases, it is not \cite{sid:sokoban,babyAI,minigrid:amigo,procgen,nethack}.  Also  in some cases, the learning is supervised; in others, it is based
on RL. Closest to our work is the use of GNNs for learning nearly-perfect general policies in planning domains using
supervised and approximate value iteration methods \cite{stahlberg:bonet:geffner:icaps:2022,stahlberg:bonet:geffner:kr:2022}

\subsubsection{General Policies Using Logical Methods} Learning general  policies has  been studied in the  planning setting
\cite{srivastava08learning,hu:generalized,BelleL16,sheila:generalized2019,anders:generalized} where logical
and  combinatorial methods have been used  \cite{khardon:generalized,martin:kr2000}.
The representation of such policies as classifiers for state transition is from \citeay{frances:aaai2021}.


\subsubsection{General Policies and Causal Models} Causal  models \cite{pearl:causality} have been used
to address the problem of out-of-distribution generalization in DL \cite{bengio:high-level,bernhard:causal},
as causal relations are modular and invariant (remain true after interventions). Recent work on policy learning
has incorporated inductive biases  motivated by causal considerations \cite{pineau:causal,causal:policy,causal-abstraction}.
Planning representations are causal too, and action schemas are modular and invariant in a domain.
It is no  accident  that we learn policies that generalize  well when using the  resulting state languages.
Methods for learning the action schemas and the predicates involved have also been developed \cite{asai:fol,bonet:ecai2020,ivan:kr2021}.

%% file: sections/conclusions.tex
\section{Conclusions}

Deep learning (DL)  and deep reinforcement learning (DRL)  have caused a revolution in AI and a significant impact
outside of AI. Yet, the methods developed in DL and DRL while incredibly powerful, are not  reliable or
transparent, and the research methodology is often too much focused on experimental performance relative to baselines
and not on understanding.  The problem of generalization has been  central in DRL for many
years but the analysis is hindered by the lack of a  language to represent states, and a language
for representing classes of problems over which the policies should generalize. The setting of classical
planning gives us both, and has helped us to understand the power of  DRL methods, and
to identify and address limitations, that have little to do with DL or DRL.
Indeed, we have shown that DRL methods learn nearly-perfect general policies out of
the box in six  of the ten domains considered, and that three  domains  fail
for structural reasons:  expressive limitations of GNNs and the
optimality/generalization tradeoff. We have addressed these limitations as well
by extending the states with derived predicates and by adapting the DRL algorithms to optimize
a different cost measure. Our  approach to learning general plans has its own limitation
such as the assumption that a lifted model of the domains is known,
that we and others have addressed elsewhere.
An overall lesson is that there is no need to choose
between crisp AI symbolic models  and data-derived  deep (reinforcement) learners.
The latter can be understood as a  new powerful class of solvers
that are worth having in the toolbox.

%% file: sections/acknowledgements.tex
\section*{Acknowledgements}


The research of H. Geffner has been supported by the Alexander von Humboldt Foundation
with funds from the Federal Ministry for Education and Research. The research has also
received funding from the European Research Council (ERC), Grant agreement No.\ 885107,
and Project TAILOR, Grant agreement No.\ 952215, under EU Horizon 2020 research and innovation
programme, the Excellence Strategy of the Federal Government and the NRW L\"{a}nder, and
the Knut and Alice Wallenberg (KAW) Foundation under the WASP program. The computations
were enabled by the supercomputing resource Berzelius provided by National Supercomputer
Centre at Link\"{o}ping University and the Knut and Alice Wallenberg foundation.